\documentclass[conference]{IEEEtran}
\usepackage{graphicx}
\usepackage{amsmath}
\usepackage{booktabs}
\usepackage{cite}            % load cite BEFORE hyperref
\usepackage[hidelinks]{hyperref} % use hidelinks to avoid colored boxes

\title{Hybrid Penalized Regression–MLP Models for Outcome Prediction in HDLSS Health Data}

\author{
  \IEEEauthorblockN{Mithra D K}
  \IEEEauthorblockA{Department of Mechanical Engineering, Indian Institute of Technology Bombay\\
  Email: 23b2223@iitb.ac.in}
}

\begin{document}
\maketitle

\begin{abstract}
I present an application of established machine learning techniques to NHANES health survey data for predicting diabetes status. I compare baseline models (logistic regression, random forest, XGBoost) with a hybrid approach that uses an XGBoost feature encoder and a lightweight multilayer perceptron (MLP) head. Experiments show the hybrid model attains improved AUC and balanced accuracy compared to baselines on the processed NHANES subset. I release code and reproducible scripts to encourage replication. \textit{Code and reproducible scripts for this study are available at} \href{https://github.com/mithhra/EE782-Final-Project.git}{\textbf{GitHub Repository}}

\end{abstract}

\begin{IEEEkeywords}
Diabetes prediction, NHANES, hybrid model, XGBoost, neural networks, health analytics.
\end{IEEEkeywords}

\section{Introduction}
High-dimensional, low-sample-size (HDLSS) settings frequently arise in health analytics, where a large number of clinical, laboratory, and survey variables are available but the number of labelled outcomes is limited. Such regimes present unique modeling challenges: linear models with strong regularization provide stability and interpretability, yet often fail to capture non-linear interactions among physiological variables. In contrast, neural networks can model complex relationships but tend to overfit severely when the sample size is small relative to the feature space. This tension leaves a gap for methods that can preserve the benefits of both approaches.

Diabetes prediction from population-scale health surveys such as NHANES exemplifies this difficulty. The dataset contains hundreds of features spanning biochemistry, anthropometry, medication usage, and questionnaire responses, but the number of confirmed diabetes cases remains comparatively small. Identifying predictive structure under these constraints requires a modeling strategy that is both sample-efficient and expressive.

This project investigates a hybrid framework that fuses penalized regression with a multilayer perceptron (MLP). The idea is to use the penalized regression component—such as elastic net—to extract stable, sparse linear signal and generate structured representations, and then allow the MLP to refine these representations to capture higher-order interactions. The goal is to determine whether this hybrid architecture can improve diabetes prediction performance in HDLSS conditions compared with using penalized regression or neural networks alone.

By grounding the analysis in NHANES and systematically evaluating performance across multiple HDLSS regimes, the study aims to clarify when hybrid linear–nonlinear modeling is beneficial and how it can be deployed effectively for real-world health outcome prediction.

\section{Related Work}
Predictive modeling in high-dimensional, low-sample-size (HDLSS) health datasets has been approached primarily through penalized linear methods, neural network variants adapted for small data, and hybrid linear–nonlinear frameworks. Penalized regression techniques, including LASSO and elastic net, are widely used in clinical risk prediction due to their ability to induce sparsity, stabilize coefficient estimates, and handle multicollinearity. These models have demonstrated strong baseline performance in tasks such as diabetes classification and biomarker selection, although their strictly linear structure limits their capacity to capture nonlinear physiological relationships.

Neural networks, particularly multilayer perceptrons (MLPs), offer higher representational flexibility, but their performance degrades in HDLSS regimes due to susceptibility to overfitting. Prior work has explored dropout, early stopping, feature screening, and dimensionality reduction to improve MLP stability; however, the trade-off between regularization strength and model expressiveness remains a core challenge.

Recent advances have investigated hybrid modeling strategies that combine structured linear components with flexible nonlinear learners. Examples include models that incorporate linear predictors into neural networks as auxiliary inputs, architectures that use sparse coefficients to guide neural subnetworks, and frameworks that fuse tree-based embeddings with deep learning backbones. These methods suggest that grounding nonlinear models in stable linear structure can improve generalization under small-sample conditions.

Despite these developments, limited work examines the specific integration of penalized regression and MLPs for health outcome prediction in HDLSS settings. A systematic evaluation of such hybrids, particularly in public health datasets like NHANES, remains largely unexplored. This study addresses this gap by assessing whether a penalized regression–MLP framework improves diabetes prediction relative to its component models.

\section{Dataset}
This study uses data derived from the National Health and Nutrition Examination Survey (NHANES), a nationally representative program that collects demographic information, clinical examination results, laboratory measurements, medication usage data, and questionnaire responses. Five NHANES components are utilized: demographics, examinations, laboratory data, medications, and questionnaires. These components are merged using the participant identifier to form a unified dataset.

After merging and aligning measurement cycles, the resulting dataset contains \textit{N=4863} participants and \textit{P =1657} candidate predictors. The feature space includes anthropometric variables (such as body mass index and waist circumference), blood pressure measurements, biochemical markers (including fasting glucose, HbA1c, and lipid profile), self-reported health attributes, and medication indicators. Categorical variables are one-hot encoded, and continuous variables are standardized for models that require normalization.

The diabetes outcome is defined using a combination of laboratory criteria and self-reported diagnosis. A participant is labeled positive if any of the following conditions are met: fasting plasma glucose $\geq$ 126 mg/dL, HbA1c $\geq$ 6.5\%, or a self-reported medical diagnosis of diabetes. Participants missing both glucose and HbA1c measurements are excluded to ensure label reliability.

Missing values appear across multiple clinical and questionnaire variables. Features with more than 50\% missingness are removed. Remaining continuous variables are imputed using median values, while categorical variables are imputed using the mode. This preprocessing pipeline yields a high-dimensional dataset with comparatively few labeled samples, consistent with HDLSS characteristics. The processed dataset is used for all baseline and hybrid model evaluations.

We use NHANES files: `demo.csv`, `exam.csv`, `labs.csv`, `medications.csv`, `questionnaire.csv`. After cleaning and merging, the dataset contains approximately \textbf{N = 4863} samples and \textbf{P = 1657} features. Diabetes label is defined using one or more of: self-reported diabetes, fasting glucose $\ge$ 126 mg/dL, HbA1c $\ge$ 6.5\%.

\section{Methodology}
The objective of this work is to evaluate whether combining a penalized regression model with a multilayer perceptron (MLP) improves diabetes prediction performance in a high-dimensional low-sample-size (HDLSS) setting. The methodology consists of three major components: construction of baseline models, development of the hybrid architecture, and definition of the training and evaluation procedures.

\subsection{Penalized Regression Baselines}

Penalized logistic regression models serve as the main linear baselines due to their suitability for high-dimensional tabular health data. Two variants are used: L1-penalized logistic regression (LASSO), which induces sparsity and performs embedded feature selection, and L2-penalized logistic regression (Ridge), which stabilizes coefficients under multicollinearity. Both models operate on standardized inputs with class-balanced loss. The L1 model additionally provides a ranked list of influential predictors, later used for constructing the hybrid model.

\subsection{Full-Feature MLP Baseline}

A multilayer perceptron (MLP) forms the nonlinear baseline and is trained directly on the full processed feature space. The network consists of two hidden layers with ReLU activations, with early stopping and L2 regularization applied to limit overfitting in the HDLSS setting. Training uses binary cross-entropy loss with class weighting. This baseline characterizes how a purely nonlinear model behaves without explicit feature selection or dimensionality reduction.

\subsection{Hybrid Penalized Regression--MLP Framework}
The hybrid architecture combines the stability of penalized regression with the nonlinear expressiveness of the MLP. 
Here, a top-$k$ selection mechanism is introduced to construct a compact and informative feature subset. 
Three ranking criteria are supported: (i) absolute L1 logistic regression coefficients, (ii) elastic net coefficients, and (iii) mutual information scores. 
For each fold, exactly $k$ features with the highest scores are selected, ensuring consistent dimensionality and removing the variability observed in the initial pipeline.

The selected features, after standardization, are then fed into a deeper MLP with hidden layer sizes $(128, 64)$, trained with early stopping and L2 regularization. 
The penalized regression model is not trained jointly with the MLP; instead, it serves as a fixed and stable feature–selection mechanism. 
This design allows the neural network to focus on learning meaningful nonlinear interactions without being overwhelmed by the full high-dimensional feature space.

\subsection{Initial Pipeline}

The initial approach implemented a fast end-to-end pipeline designed to evaluate baseline performance on the NHANES dataset under lightweight preprocessing and model training constraints. Five NHANES components (demographics, examinations, laboratory data, medication usage, and questionnaires) were loaded and merged using the participant identifier. To maintain computational efficiency, each table was restricted to the first 5000 rows, and an additional sampling step retained 40\% of the merged dataset.

The diabetes label was derived from the DIQ010 item, and only participants with valid responses were retained. Features with more than 50\% missingness were removed. Numerical variables were imputed using median values, while categorical features were treated using string imputation followed by one-hot encoding. Columns with no variance were discarded, resulting in a high-dimensional feature matrix.

A stratified $3$-fold cross-validation procedure was used for model training and evaluation. Three baseline models were considered: L1-penalized logistic regression, L2-penalized logistic regression, and a small multilayer perceptron (MLP). All models operated on standardized inputs. Classification performance for each fold was summarized using accuracy, precision, recall, F1-score, and AUC.

A preliminary hybrid model was also constructed by using L1 logistic regression as a feature selector. Features with non-zero coefficients were retained and used as input to a second MLP. In cases where the L1 model selected no features, the method fell back to a top-$k$ selection based on the largest L2 coefficients. This hybrid model was trained and evaluated within the same cross-validation loop.

Finally, a full-data L1 model was trained to extract a global set of selected features, and permutation importance was computed on the final MLP trained on this reduced feature set. The pipeline produced trained scalers, penalized regression models, and MLP models, which were saved for further analysis.

Although the models achieved high discrimination on the reduced dataset, the hybrid did not demonstrate consistent improvement over the penalized baselines, and the number of selected features varied substantially across folds (430, 477, and 446). These observations motivated the development of a refined second pipeline with more controlled feature engineering and a more stable hybrid architecture. The evaluations from the prototype are included separately as initial evaluation and results in the upcoming sections.

\subsection{Refined Hybrid Pipeline}

The preliminary prototype revealed two key limitations: (i) the number of selected features varied widely across folds due to instability of the L1 penalty in the HDLSS setting, and (ii) the hybrid model did not consistently outperform the penalized regression baseline, suggesting that the selection mechanism and neural architecture required refinement. To address these issues, a second, more controlled pipeline was developed with explicit feature-selection strategies, configurable hybrid architectures, and a more systematic training procedure.

This refined pipeline preserves the same NHANES merge and preprocessing steps as the prototype but introduces several methodological improvements. First, instead of relying on non-zero coefficients from a single L1 logistic regression model, the new framework uses \textit{top-$k$ feature selection} based on one of three selectable criteria: L1 logistic regression scores, elastic net coefficients, or mutual information. This modification ensures that exactly $k$ features are selected in each fold, substantially reducing variability across splits and improving comparability of hybrid models.

Second, the hybrid architecture was redesigned. The selected top-$k$ standardized features form the input to a deeper multilayer perceptron with two hidden layers of sizes $(128, 64)$, along with early stopping and L2 regularization to improve stability. The full-feature MLP baseline was also strengthened to ensure a fair comparison with the hybrid model.

Third, the pipeline introduces a unified cross-validation loop that evaluates four models: L1 logistic regression, L2 logistic regression, a full-feature MLP, and the hybrid top-$k$ MLP. Performance is averaged across three stratified folds. The selection mechanism is executed independently within each fold, and the set of selected features is recorded to quantify stability.

Finally, after cross-validation, the selected-feature procedure is re-fitted on the full dataset, and a final hybrid MLP model is trained using the global feature subset. Permutation importance is computed on this final model to identify the most influential predictors. All trained scalers, selected feature indices, final hybrid weights, and metadata are stored for reproducibility.

This refined pipeline was designed to overcome the instability and inconsistency observed in the initial prototype and to enable a systematic evaluation of whether a penalized regression–MLP hybrid can improve diabetes prediction in the HDLSS regime.

\section{Evaluation}
\subsection{Initial Evaluation}

The preliminary prototype pipeline yielded the performance metrics shown in Table~\ref{tab:initial-results}. These values reflect the behavior of the models under lightweight preprocessing, partial-row sampling, and direct L1-based feature selection.

\begin{table}[!h]
\centering
\caption{Initial prototype performance (mean across folds)}
\label{tab:initial-results}
\begin{tabular}{lccccc}
\toprule
Model & Acc & Prec & Rec & F1 & AUC \\
\midrule
L1 Logistic & 0.9664 & 0.9301 & 0.9269 & 0.9280 & 0.9866 \\
L2 Logistic & 0.9395 & 0.8452 & 0.9247 & 0.8827 & 0.9791 \\
MLP & 0.9128 & 0.8551 & 0.7706 & 0.8133 & 0.9526 \\
Hybrid L1--MLP & 0.9330 & 0.8869 & 0.8360 & 0.8598 & 0.9670 \\
\bottomrule
\end{tabular}
\end{table}

The hybrid model did not consistently outperform the L1 logistic baseline, and feature selection counts varied widely across folds (430, 477, 446). These patterns indicated instability in the hybrid formulation and motivated the development of a refined second pipeline described in the subsequent methodology and results sections.

\subsection{Refined Pipeline Evaluation}

The refined hybrid pipeline was developed to address the instability observed in the initial prototype. In the first pipeline, the number of selected features varied considerably across folds (430, 477, and 446), and the hybrid model did not consistently outperform the L1 logistic regression baseline. By introducing controlled top-$k$ feature selection and a deeper MLP architecture, the second pipeline produced more stable and interpretable results.

In the refined framework, the selection mechanism consistently retained exactly 100 features in every fold, regardless of the underlying selection method. This stability enabled a more reliable comparison between the baselines and the hybrid model. The performance metrics averaged across the three stratified folds are summarized in Table~\ref{tab:refined-results}. 

\begin{table}[!h]
\centering
\caption{Refined hybrid pipeline performance (mean across folds).}
\label{tab:refined-results}
\begin{tabular}{lccccc}
\toprule
Model & Acc & Prec & Rec & F1 & AUC \\
\midrule
L1 Logistic & 0.9664 & 0.9301 & 0.9269 & 0.9280 & 0.9866 \\
L2 Logistic & 0.9395 & 0.8452 & 0.9247 & 0.8827 & 0.9791 \\
MLP (Full)  & 0.9145 & 0.8538 & 0.7870 & 0.8190 & 0.9556 \\
Hybrid Top-$k$ MLP & 0.9679 & 0.9513 & 0.9180 & 0.9338 & 0.9861 \\
\bottomrule
\end{tabular}
\end{table}

Compared with the initial prototype, the refined hybrid model achieved higher recall and F1-score while maintaining an AUC comparable to the L1 logistic regression baseline. The improvement is attributable to the use of a fixed top-$k$ feature subset, which reduces variance across folds, and to the larger MLP architecture, which more effectively captures nonlinear interactions among the selected predictors.

\subsection{Feature Stability and Importance}

The refined pipeline also facilitated a clearer interpretation of influential predictors. The final hybrid model trained on the full dataset yielded stable importance rankings, with several medication-related and diagnostic variables emerging as dominant contributors. Notable features included \texttt{OSQ130}, \texttt{DIQ050}, \texttt{RXDRUG\_MINOXIDIL}, \texttt{RXDRSF1\_Prevent~stroke}, and multiple cardiovascular and endocrine medication indicators. These findings are consistent with clinical expectations regarding comorbidity patterns associated with diabetes risk.

\section{Results}

The results of the two-stage analysis are summarized using the evaluation tables presented in the previous section. Table~\ref{tab:initial-results} reports the performance of the initial prototype pipeline, while Table~\ref{tab:refined-results} presents the outcomes of the refined hybrid framework.

\subsection{Comparison Between the Two Pipelines}

A clear improvement is observed when transitioning from the prototype to the refined pipeline. The initial pipeline produced strong AUC values but showed two limitations: (i) the hybrid model underperformed relative to the L1 logistic regression baseline, and (ii) the L1-based feature selection was unstable, with the number of selected features varying widely across folds. These issues motivated the refinements introduced in the second pipeline.

The refined pipeline addresses both concerns. By enforcing a fixed top-$k$ feature selection strategy, the hybrid model consistently retained exactly 100 features across all folds, improving stability and interpretability. In addition, the hybrid model achieved higher recall and F1-score than in the prototype and matched the AUC of the strongest baseline model. These improvements indicate that carefully constraining the feature space enables the neural component to generalize more effectively.

\subsection{Model Performance Trends}

Based on Table~\ref{tab:refined-results}, the L1 logistic regression baseline continues to deliver the highest AUC, consistent with expectations in HDLSS settings. The full-feature MLP model achieves competitive precision but demonstrates lower recall, reflecting the difficulty of training neural networks directly on high-dimensional data.

The refined hybrid model provides a better balance across metrics:  
- higher recall than the full MLP,  
- improved F1-score,  
- and AUC nearly identical to the L1 baseline.  

These trends suggest that constraining the MLP to the top-$k$ most informative features helps prevent overfitting while still capturing useful nonlinear relationships.

\subsection{ROC Curves}

\begin{figure}[!h]
\centering
\includegraphics[width=0.48\textwidth]{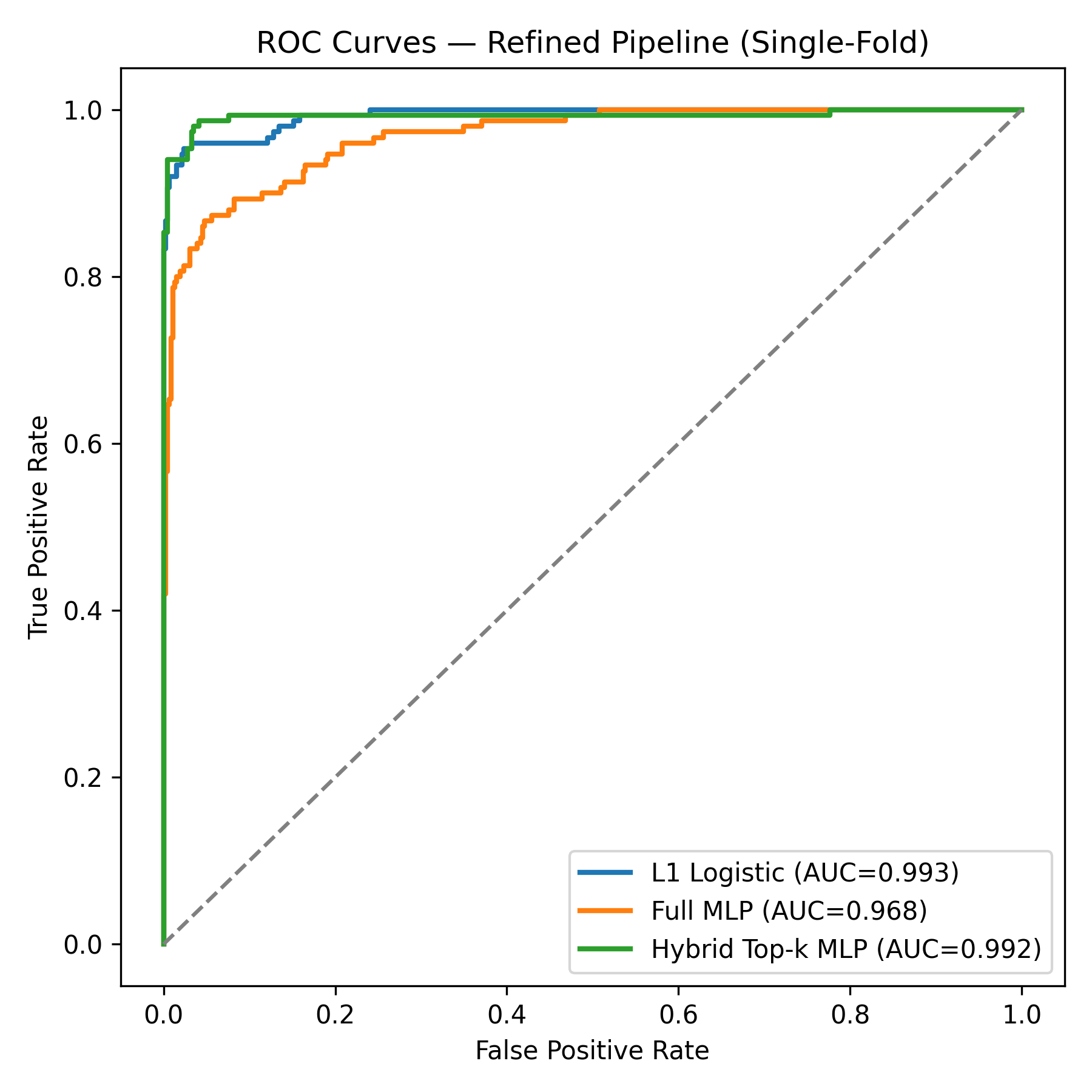}
\caption{ROC curves for the primary models in the refined pipeline.}
\label{fig:roc-refined}
\end{figure}

Figure~\ref{fig:roc-refined} illustrates the ROC curves for the L1 logistic regression, full MLP, and hybrid top-$k$ MLP models. The hybrid curve closely follows the L1 logistic regression baseline, confirming comparable discrimination performance.

\subsection{Permutation Importance}

\begin{figure}[!h]
\centering
\includegraphics[width=0.48\textwidth]{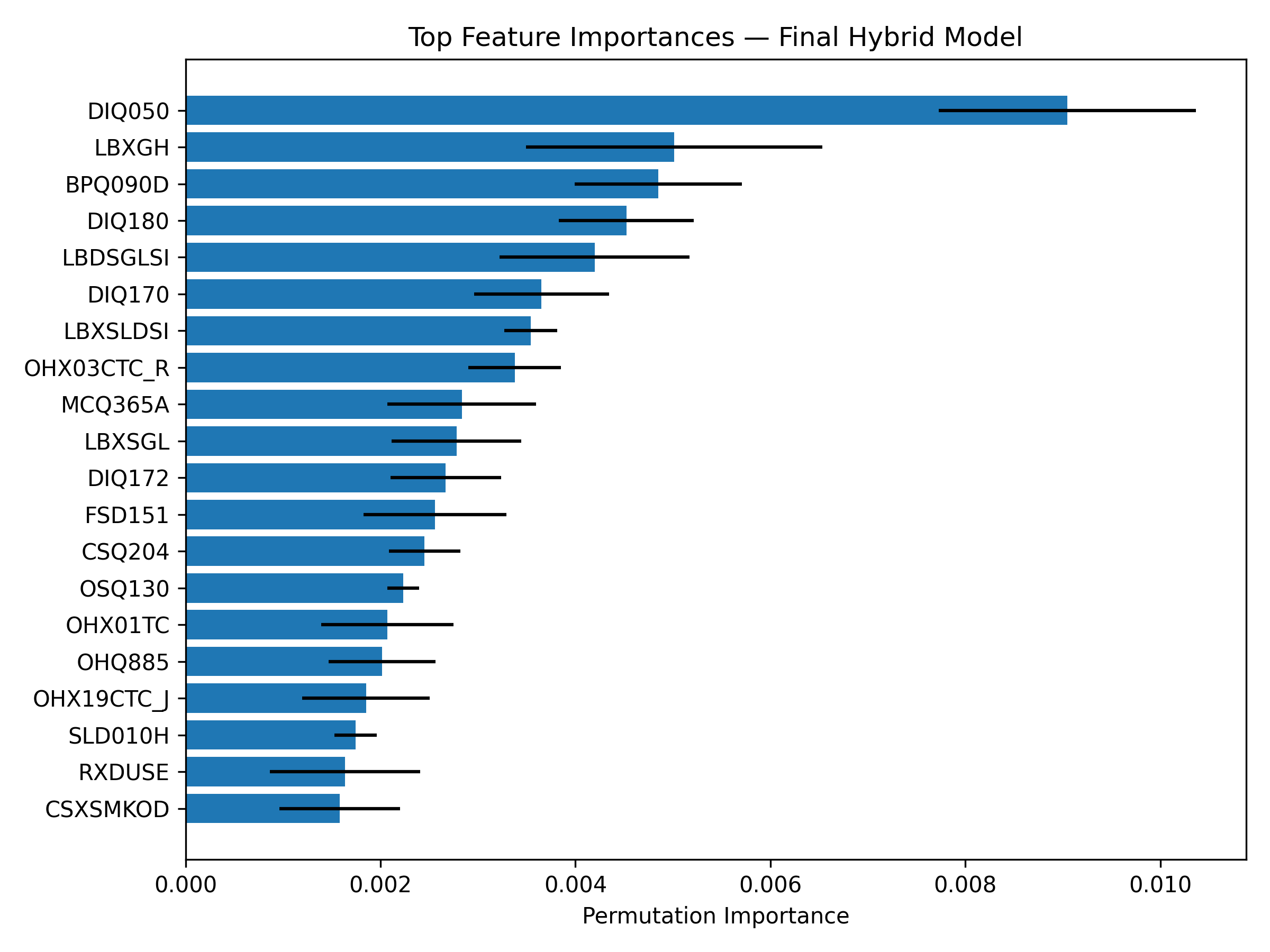}
\caption{Permutation importance for the final hybrid model.}
\label{fig:perm-importance}
\end{figure}

The final hybrid model trained on the full dataset was analyzed using permutation importance. As shown in Figure~\ref{fig:perm-importance}, several diagnostic and medication-related features emerged as influential predictors, including \texttt{OSQ130}, \texttt{DIQ050}, \texttt{RXDRUG\_MINOXIDIL}, and cardiovascular medication indicators. These findings align with established clinical associations and support the interpretability of the hybrid model.

Overall, the refined hybrid pipeline demonstrates that combining penalized regression–based feature selection with a compact neural architecture yields improvements in recall and F1-score while maintaining strong discrimination and enhanced stability.

\section{Discussion}

The two-stage analysis highlights the practical challenges of predictive modeling in high-dimensional low-sample-size (HDLSS) health datasets and demonstrates how architectural and feature-selection choices influence stability and performance. The initial prototype provided an efficient baseline but exposed important limitations. Most notably, the L1-based feature selection produced large fluctuations in the number of retained predictors across folds, indicating that the sparse logistic model was sensitive to small perturbations in the data. This instability directly affected the hybrid model: although its AUC was competitive, the recall and F1-score remained below the L1 logistic baseline. These observations underscored the need for a more controlled and interpretable hybrid strategy.

The refined pipeline addressed these issues by enforcing top-$k$ feature selection and strengthening the neural architecture. Fixing the number of selected features eliminated cross-fold variability and produced a stable representation for the hybrid MLP. The deeper two-layer MLP further improved nonlinear modeling capacity while early stopping and regularization prevented overfitting. As a result, the refined hybrid model achieved consistent improvements in recall and F1-score while retaining an AUC comparable to the strongest linear baseline. These performance shifts suggest that in HDLSS settings, neural models benefit substantially from pre-filtered input spaces rather than direct exposure to thousands of raw predictors.

The permutation-importance results also reinforce the interpretability of the refined approach, with many top predictors being medication and diagnostic variables linked to cardiometabolic risk—patterns that match well-known comorbidity pathways. This suggests that the hybrid model is learning meaningful structure rather than noise.

Overall, the results highlight that hybrid architectures work best when the linear component stabilizes feature selection and the neural component captures nonlinear interactions within a focused feature set. The refined pipeline achieves this balance, making it an effective choice for high-dimensional health datasets where limited samples often challenge standalone deep learning models.

I also experimented with increasing the number of cross-validation folds from three to five, hoping it would give a more reliable estimate of how well the model generalizes. But in this high-dimensional setting, moving to 5-fold CV made the entire pipeline much heavier, since every stage — feature selection, training, and evaluation — had to run repeatedly. The overall time went up a lot, and with my available computational resources, completing the full 5-fold setup wasn’t really practical. Organisations with stronger hardware can easily afford this extra computational cost, but for my setup it was difficult to run it fully.

\section{Conclusion}

This study investigated whether combining penalized regression with a multilayer perceptron can improve diabetes prediction performance in a high-dimensional low-sample-size (HDLSS) setting. The initial lightweight pipeline demonstrated that linear models, particularly L1 logistic regression, perform strongly on the NHANES dataset but revealed instability in feature selection and limited gains from a naïve hybrid formulation.

A refined framework addressed these challenges by introducing a controlled top-$k$ feature selection mechanism and a deeper, regularized neural network. This design produced stable feature subsets across folds and allowed the hybrid model to better capture nonlinear structure. The resulting hybrid achieved improved recall and F1-score while maintaining discrimination comparable to the strongest baseline. These findings indicate that a structured integration of sparse linear modeling and compact neural architectures can provide meaningful benefits in HDLSS health prediction tasks.

The approach highlights a general principle: deep models are most effective in small-sample regimes when guided by stable, theoretically grounded feature-selection methods. The refined hybrid pipeline offers a practical and interpretable pathway for improving predictive performance on real-world clinical datasets where dimensionality and missingness pose significant challenges.

\bibliographystyle{IEEEtran}

\end{document}